\documentclass[runningheads,a4paper]{llncs}[2015/06/24]

\usepackage{cmap}
\usepackage[T1]{fontenc}

\usepackage{graphicx}

\usepackage[ngerman,english]{babel}
\addto\extrasenglish{\languageshorthands{ngerman}\useshorthands{"}}

\usepackage[%
rm={oldstyle=false,proportional=true},%
sf={oldstyle=false,proportional=true},%
tt={oldstyle=false,proportional=true,variable=true},%
qt=false%
]{cfr-lm}
\usepackage[math]{blindtext}

\usepackage{cite}

\usepackage{paralist}

\usepackage{csquotes}

\usepackage{microtype}

\usepackage{subcaption}

\usepackage{url}
\makeatletter
\g@addto@macro{\UrlBreaks}{\UrlOrds}
\makeatother

\usepackage{xcolor}

\usepackage{pdfcomment}
\hypersetup{hidelinks,
   colorlinks=true,
   allcolors=black,
   pdfstartview=Fit,
   breaklinks=true}
\usepackage[all]{hypcap}

\usepackage[capitalise,nameinlink]{cleveref}
\crefname{section}{Sect.}{Sect.}
\Crefname{section}{Section}{Sections}

\usepackage{xspace}

\usepackage{bm}
\usepackage{amsmath}
\usepackage{mathtools}

\DeclareFontFamily{U}{MnSymbolC}{}
\DeclareSymbolFont{MnSyC}{U}{MnSymbolC}{m}{n}
\DeclareFontShape{U}{MnSymbolC}{m}{n}{
    <-6>  MnSymbolC5
   <6-7>  MnSymbolC6
   <7-8>  MnSymbolC7
   <8-9>  MnSymbolC8
   <9-10> MnSymbolC9
  <10-12> MnSymbolC10
  <12->   MnSymbolC12%
}{}
\DeclareMathSymbol{\powerset}{\mathord}{MnSyC}{180}

\begin{document}

\title{Deep Convolutional Neural Networks for Breast Cancer Histology Image Analysis}
\titlerunning{Deep Learning for Breast Cancer Histology}


%
\author{Alexander Rakhlin\inst{1}, Alexey Shvets\inst{2}, Vladimir Iglovikov\inst{3},\\ \and Alexandr A. Kalinin\inst{4} }
\authorrunning{Rakhlin et al.}

\institute{
Neuromation, St. Petersburg 191025, Russia\\
\email{rakhlin@neuromation.io} \and Massachusetts Institute of Technology, Cambridge, MA 02142, USA\\
\email{shvets@mit.edu} \and Lyft Inc., San Francisco, CA 94107, USA\\ \email{iglovikov@gmail.com}  \and University of Michigan, Ann Arbor, MI 48109, USA\\ \email{akalinin@umich.edu}
}
\maketitle

\begin{abstract}
Breast cancer is one of the main causes of cancer death worldwide. Early diagnostics significantly increases the chances of correct treatment and survival, but this process is tedious and often leads to a disagreement between pathologists. Computer-aided diagnosis systems showed potential for improving the diagnostic accuracy. In this work, we develop the computational approach based on deep convolution neural networks for breast cancer histology image classification. Hematoxylin and eosin stained breast histology microscopy image dataset is provided as a part of the ICIAR 2018 Grand Challenge on Breast Cancer Histology Images. Our approach utilizes several deep neural network architectures and gradient boosted trees classifier. For 4-class classification task, we report 87.2\% accuracy. For 2-class classification task to detect carcinomas we report 93.8\% accuracy, AUC 97.3\%, and sensitivity/specificity 96.5/88.0\% at the high-sensitivity operating point. To our knowledge, this approach outperforms other common methods in automated histopathological image classification. The source code for our approach is made publicly available at \url{https://github.com/alexander-rakhlin/ICIAR2018}
\end{abstract}

\begin{keywords}
Medical imaging, Computer-aided diagnosis (CAD), Computer vision, Image recognition, Deep learning
\end{keywords}

\section{Introduction}\label{sec:intro}
Breast cancer is the most common cancer diagnosed among US women (excluding skin cancers), accounting for 30\% of all new cancer diagnoses in women in the United States \cite{siegel2018stats}. Breast tissue biopsies allow the pathologists to histologically assess the microscopic structure and elements of the tissue. Histopathology aims to distinguish between normal tissue, non-malignant (benign) and malignant lesions (carcinomas) and to perform a prognostic evaluation \cite{elston1991pathological}. A combination of hematoxylin and eosin (H\&E) is the principal stain of tissue specimens for routine histopathological diagnostics. There are multiple types of breast carcinomas that embody characteristic tissue morphology, see Fig. \ref{fig::classes}. Breast carcinomas arise from the mammary epithelium and cause a pre-malignant epithelial proliferation within the ducts, called ductal carcinoma \textit{in situ}. Invasive carcinoma is characterized by the cancer cells gaining the capacity to break through the basal membrane of the duct walls and infiltrate into surrounding tissues \cite{ROBERTSON2017digital}.

Morphology of tissue, cells, and subcellular compartments is regulated by complex biological mechanisms related to cell differentiation, development, and cancer \cite{kalinin20173d}. Traditionally, morphological assessment and tumor grading were visually performed by the pathologist, however, this process is tedious and subjective, causing inter-observer variations even among senior pathologists \cite{meyer2005breast, elmore2015diagnostic}. The subjectivity of the application of morphological criteria in visual classification motivates the use of computer-aided diagnosis (CAD) systems to improve the diagnosis accuracy, reduce human error, increase the level of inter-observer agreement, and increased reproducibility \cite{ROBERTSON2017digital}.

There are many methods developed for the digital pathology image analysis, from rule-based to applications of machine learning \cite{ROBERTSON2017digital}. Recently, deep learning based approaches were shown to outperform conventional machine learning methods in many image analysis task, automating end-to-end processing \cite{ching2017opportunities, iglovikov2017satellite, iglovikov2018ternausnet}. In the domain of medical imaging, convolutional neural networks (CNN) have been successfully used for diabetic retinopathy screening \cite{rakhlin2017diabetic}, bone disease prediction \cite{tiulpin2018automatic} and age assessment \cite{iglovikov2017pediatric}, and other problems \cite{ching2017opportunities}. Previous deep learning-based applications in histological microscopic image analysis have demonstrated their potential to provide utility in diagnosing breast cancer \cite{spanhol2016breast, araujo2017classification, bejnordi2017diagnostic, ROBERTSON2017digital}.

In this paper, we present an approach for histology microscopy image analysis for breast cancer type classification. Our approach utilizes deep CNNs for feature extraction and gradient boosted trees for classification and, to our knowledge, outperforms other similar solutions.

\section{Methods}
\subsection{Dataset}
\begin{figure}[!t]
\centering
\includegraphics[width=\linewidth]{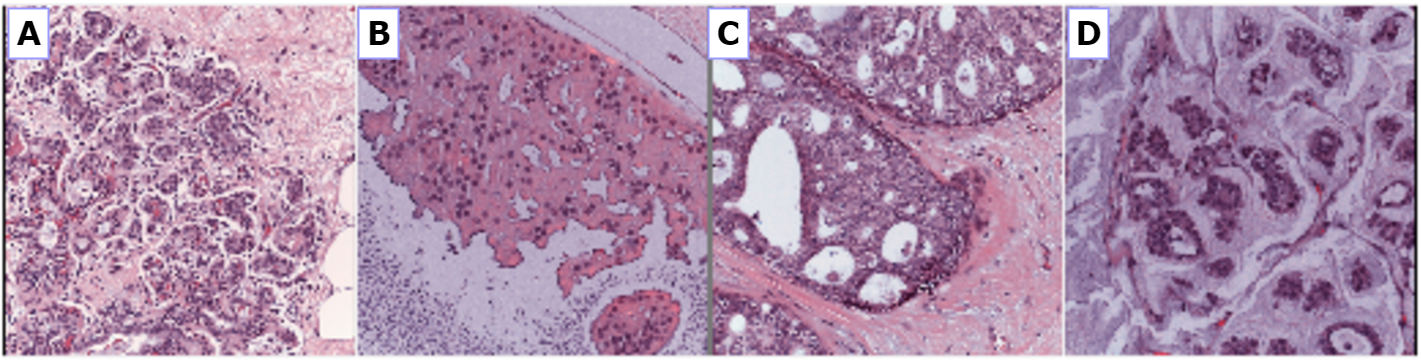}
\caption{Examples of microscopic biopsy images in the dataset: (A) normal; (B) benign; (C) \textit{in situ} carcinoma; and (D) invasive carcinoma}
\label{fig::classes}
\end{figure}

The image dataset is an extension of the dataset from \cite{araujo2017classification} and consists of 400 H\&E stain images ($2048\times1536$ pixels). All the images are digitized with the same acquisition conditions, with a magnification of $200\times$ and pixel size of $0.42\mu m\times0.42\mu m$. Each image is labeled with one of the four balanced classes: normal, benign, \textit{in situ} carcinoma, and invasive carcinoma, where class is defined as a predominant cancer type in the image, see Fig. \ref{fig::classes}. The image-wise annotation was performed by two medical experts \cite{iciar2018}. The goal of the challenge is to provide an automatic classification of each input image.

\subsection{Approach overview}
The limited size of the dataset (400 images of 4 classes) poses a significant challenge for the training of a deep learning model \cite{ching2017opportunities}. Very deep CNN architectures that contain millions of parameters such as VGG, Inception and ResNet have achieved the state-of-the-art results in many computer vision tasks \cite{szegedy2017inception}. However, training these neural networks from scratch requires a large number of images, as training on a small dataset leads to overfitting i.e. inability to generalize knowledge. A typical remedy in these circumstances is called fine-tuning when only a part of the pre-trained neural network is being fitted to a new dataset. However, in our experiments, fine-tuning approach did not demonstrate good performance on this task. Therefore, we employed a different approach known as deep convolutional feature representation \cite{guo2016deep}. To this end, deep CNNs, trained on large and general datasets like ImageNet (10M images, 20K classes) \cite{deng2009imagenet}, are used for unsupervised feature representation extraction. In this study, breast histology images are encoded with the state-of-the-art, general purpose networks to obtain sparse descriptors of low dimensionality (1408 or 2048). This unsupervised dimensionality reduction step significantly reduces the risk of overfitting on the next stage of supervised learning.

We use LightGBM as a fast, distributed, high performance implementation of gradient boosted trees for supervised classification \cite{ke2017lightgbm}. Gradient boosting models are being extensively used in machine learning due to their speed, accuracy, and robustness against overfitting \cite{natekin2013gradient}.

\subsection{Data pre-processing and augmentation}
To bring the microscopy images into a common space to enable improved quantitative analysis, we normalize the amount of H\&E stained on the tissue as described in \cite{macenko2009method}. For each image, we perform 50 random color augmentations. Following \cite{ruifrok2001quantification} the amount of H\&E is adjusted by decomposing the RGB color of the tissue into H\&E color space, followed by multiplying the magnitude of H\&E of every pixel by two random uniform variables from the range $[0.7, 1.3]$. Furthermore, in our initial experiments, we used different image scales, the original $2048\times1536$ pixels and downscaled in half to $1024\times768$ pixels. From the images of the original size we extract random crops of two sizes $800\times800$ and $1300\times1300$. From the downscaled images we extract crops of $400\times400$ pixels and $650\times650$ pixels. Lately, we found downscaled images is enough. Thereby, each image is represented by 20 crops. The crops are then encoded into 20 descriptors. Then, the set of 20 descriptors is combined through 3-norm pooling \cite{boureau2010theoretical} into a single descriptor:
\begin{equation}
    \mathbf{d}_{pool}=\left(\frac{1}{N}\sum\limits_{i=1}^N(\mathbf{d}_i)^p\right)^{\frac{1}{p}},
\end{equation}
where the hyperparameter $p=3$ as suggested in \cite{boureau2010theoretical, xu2015deep}, $N$ is the number of crops, $\mathbf{d}_i$ is descriptor of a crop and $\mathbf{d}_{pool}$ is pooled descriptor of the image. The p-norm of a vector gives the average for $p = 1$ and the max for  $p\rightarrow\infty$. As a result, for each original image, we obtain $50$ (number of color augmentations) $\times2$ (crop sizes) $\times3$ (CNN encoders) $=300$ descriptors.

\begin{figure}[!t]
\centering
\includegraphics[width=\linewidth]{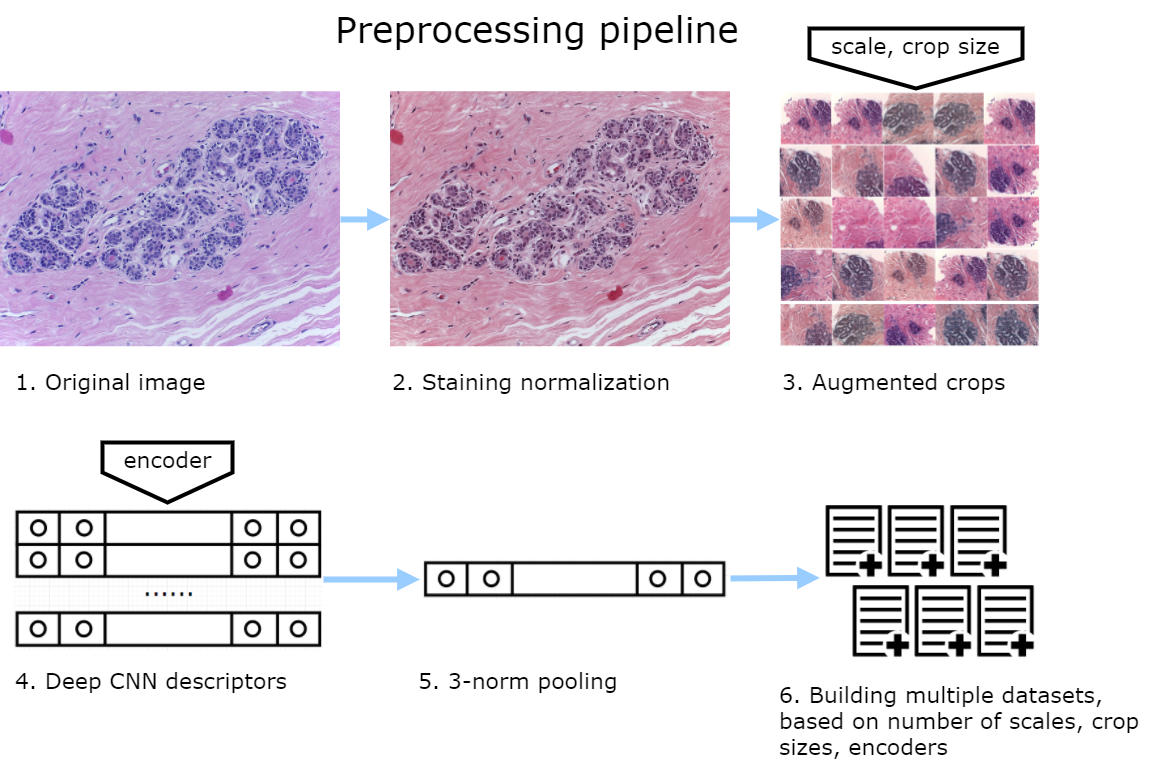}
\caption{An overview of the pre-processing pipeline.}
\label{fig::pipeline}
\end{figure}

\subsection{Feature extraction}
\begin{figure}[!t]
\centering
\includegraphics[width=\linewidth]{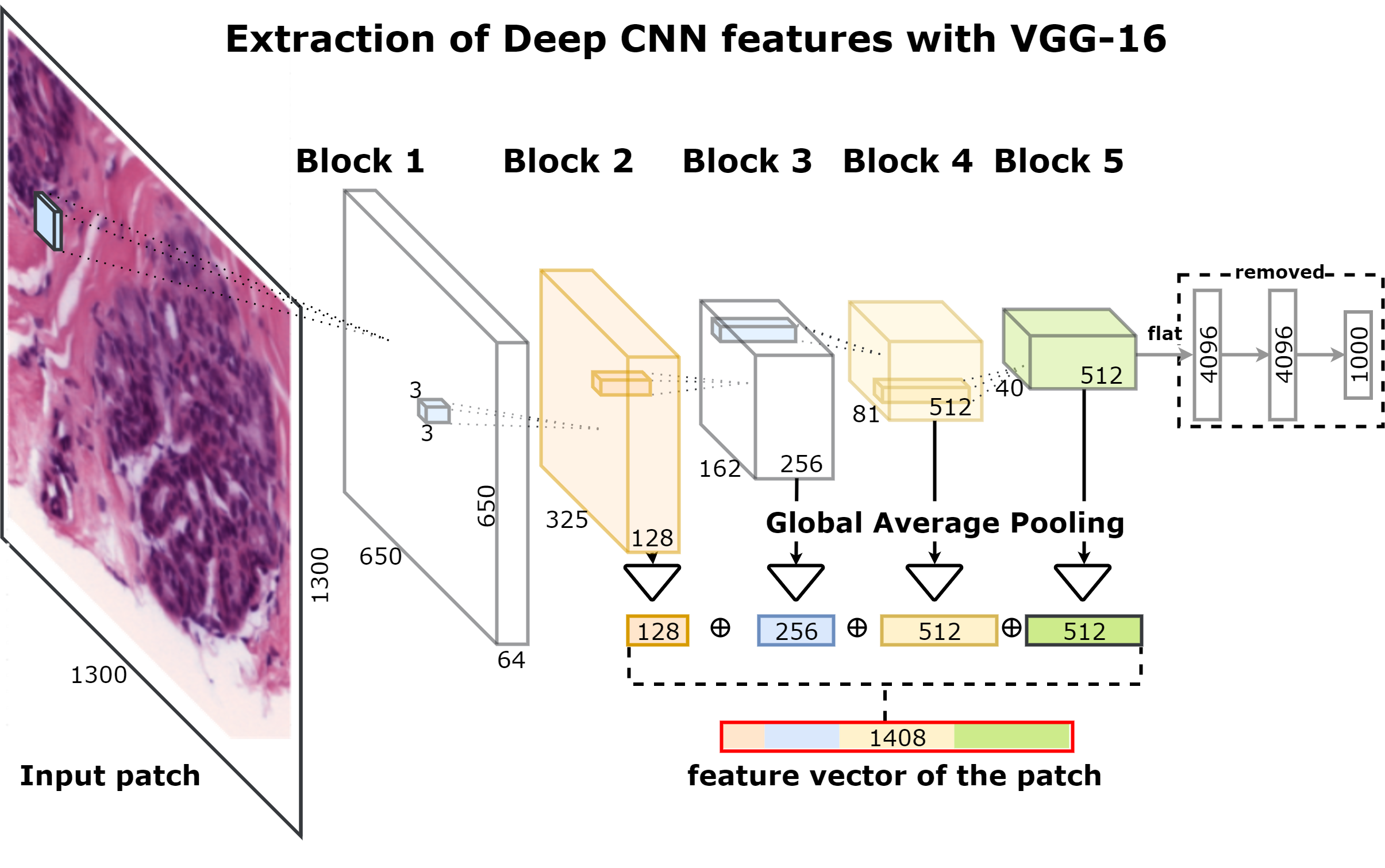}
\caption{Schematic overview of the network architecture for deep feature extraction.}
\label{fig::vgg}
\end{figure}

Overall pre-processing pipeline is depicted in Fig. \ref{fig::pipeline}. For features extraction, we use standard pre-trained ResNet-50, InceptionV3 and VGG-16 networks from Keras distribution \cite{chollet2015keras}. We remove fully connected layers from each model to allow the networks to consume images of an arbitrary size. In ResNet-50 and InceptionV3, we convert the last convolutional layer consisting of 2048 channels via \texttt{GlobalAveragePooling} into a one-dimensional feature vector with a length of 2048. With VGG-16 we apply the \texttt{GlobalAveragePooling} operation to the four internal convolutional layers: \texttt{block2}, \texttt{block3}, \texttt{block4}, \texttt{block5} with 128, 256, 512, 512 channels respectively. We concatenate them into one vector with a length of 1408, see Fig. \ref{fig::vgg}. 

\subsection{Training}
We split the data into 10 stratified folds to preserve class distribution. Augmentations increase the size of the dataset $\times300$ (2 patch sizes x 3 encoders x 50 color/affine augmentations). Nevertheless, the descriptors of a given image remain correlated. To prevent information leakage, all descriptors of the same image must be contained in the same fold. For each combination of the encoder, crop size and scale we train 10 gradient boosting models with 10-fold cross-validation. In addition to obtaining cross-validated results, this allows us to increase the diversity of the models with limited data (bagging). Furthermore, we recycle each dataset 5 times with different random seeds in LightGBM adding augmentation on the model level. As a result, we train $10$ (number of folds) $\times5$ (seeds) $\times4$ (scale and crop) $\times3$ (CNN encoders) $=600$ gradient boosting models. At the cross-validation stage, we predict every fold only with the models not trained on this fold. For the test data, we similarly extract 300 descriptors for each image and use them with all models trained for particular patch size and encoder. The predictions are averaged over all augmentations and models. Finally, the predicted class is defined by the maximum probability score.

\begin{figure}[!h]
\centering
\begin{subfigure}{.5\textwidth}
  \centering
  \includegraphics[width=\linewidth]{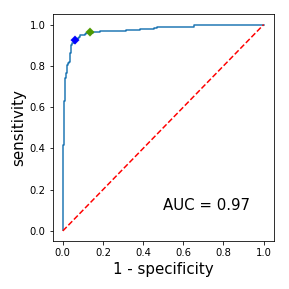}
  \label{fig:sub1}
\end{subfigure}%
\begin{subfigure}{.5\textwidth}
  \centering
  \includegraphics[width=\linewidth]{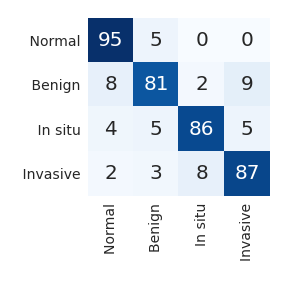}
  \label{fig:sub2}
\end{subfigure}
\caption[]{a) Non-carcinoma vs. carcinoma classification, ROC. High sensitivity setpoint=0.33 (green): 96.5\% sensitivity and 88.0\% specificity to detect carcinomas. Setpoint=0.50 (blue): 93.0\% sensitivity and 94.5\% specificity b) Confusion matrix, without normalization. Vertical axis - ground truth, horizontal - predictions.}  
\label{fig::cm_roc}
\end{figure}


\begin{table}[h!]
\caption{Accuracy (\%) and standard deviation for 4-class classification evaluated over 10 folds via cross-validation. Results for the blended model is in the bottom. Model name represented as $\langle$CNN$\rangle$-$\langle$crop size$\rangle$,  thereby VGG-650 denotes LightGBM trained on deep features extracted from 650x650 crops with VGG-16 encoder. Each column in the table corresponds to the fold number.}
\label{table:folds}
\centering   
\begin{tabular}{lllllllllllll}
\hline\noalign{\smallskip}
 & f1 & f2 & f3 & f4 & f5 & f6 & f7 & f8 & f9 & f10 & \textbf{mean} & \textbf{std}  \\
\hline
\noalign{\smallskip}
ResNet-400 & 92.0 & 77.5 & 86.5 & 87.5 & 79.5 & 84.0 & 85.0 & 83.0 & 84.0 & 82.5 & 84.2 & 4.2 \\
ResNet-650 & 91.0 & 77.5 & 86.0 & 89.5 & 81.0 & 74.0 & 85.5 & 83.0 & 84.5 & 82.5 & 83.5 & 5.2 \\
VGG-400 & 87.5 & 83.0 & 81.5 & 84.0 & 84.0 & 82.5 & 80.5 & 82.0 & 87.5 & 83.0 & 83.6 & 2.9 \\
VGG-650 & 89.5 & 85.5 & 78.5 & 85.0 & 81.0 & 78.0 & 81.5 & 85.5 & 89.0 & 80.5 & 83.4 & 4.4 \\
Inception-400 & 93.0 & 86.0 & 71.5 & 92.0 & 85.0 & 84.5 & 82.5 & 79.0 & 79.5 & 76.5 & 83.0 & 6.5 \\
Inception-650 & 91.0 & 84.5 & 73.5 & 90.0 & 84.0 & 81.0 & 82.0 & 84.5 & 78.0 & 77.0 & 82.5 & 5.5 \\
\hline
std & 1.8 & 3.5 & 5.7 & 2.8 & 2.0 & 3.7 & 1.8 & 2.1 & 3.9 & 2.7 & 3.0 & - \\
\hline
Model fusion & 92.5 & 82.5 & \bf{87.5} & 87.5 & \bf{87.5} & \bf{90.0} & 85.0 & \bf{87.5} & 87.5 & \bf{85.0} & \bf{87.2} & \bf{2.6} \\
\hline
\end{tabular}
\end{table}

\section{Results}
To validate the approach we use 10-fold stratified cross-validation.

For 2-class non-carcinomas (normal and benign) vs. carcinomas (\textit{in situ} and invasive) classification accuracy was 93.8$\pm$2.3$\%$, the area under the ROC curve was 0.973, see Fig.\ref{fig::cm_roc}a. At high sensitivity setpoint 0.33 the sensitivity of the model to detect carcinomas was 96.5$\%$ and specificity 88.0$\%$. At the setpoint 0.50 the sensitivity of the model was 93.0$\%$ and specificity 94.5$\%$, Fig. \ref{fig::cm_roc}a. Out of 200 carcinomas cases only 9 \textit{in situ} and 5 invasive were missed, Fig.\ref{fig::cm_roc}b.

Table \ref{table:folds} shows classification accuracy for 4-class classification. Accuracy averaged across all folds was 87.2$\pm$2.6$\%$. Finally, the importance of strong augmentation and model fusion we use is particularly evident from the Table \ref{table:folds}. The fused model accuracy is by 4-5\% higher than any of its individual constituents. The standard deviation of the ensemble across 10 folds is twice as low than the average standard deviation of the individual models. Moreover, all our results in the Table \ref{table:folds} are slightly improved by averaging across 5 seeded models.

\section{Conclusions}
In this paper, we propose a simple and effective method for the classification of H$\&$E stained histological breast cancer images in the situation of very small training data (few hundred samples). To increase the robustness of the classifier we use strong data augmentation and deep convolutional features extracted at different scales with publicly available CNNs pretrained on ImageNet. On top of it, we apply highly accurate and prone to overfitting implementation of the gradient boosting algorithm. Unlike some previous works, we purposely avoid training neural networks on this amount of data to prevent suboptimal generalization.

To our knowledge, the reported results are superior to the automated analysis of breast cancer images reported in literature \cite{spanhol2016breast, araujo2017classification, bejnordi2017diagnostic}.

\subsubsection*{Acknowledgments}
The authors thank the Open Data Science community \cite{ods} for useful suggestions and other help aiding the development of this work.

\bibliographystyle{splncs03}



\end{document}